\documentclass{article}
\usepackage{spconf,amsmath,graphicx,amssymb}
\usepackage{comment}
\usepackage[ruled,vlined]{algorithm2e}
\usepackage[table,xcdraw]{xcolor}

\def\R#1{(\ref{#1})}

\newcommand*{\mb}[1]{{\mathbf{#1}}}

\newcommand*{\bbR}{\mathbb{R}}

\title{Improved and efficient inter-vehicle distance estimation using road gradients of both ego and target vehicles}

\name{Muhyun Back$^1$, Jinkyu Lee$^1$, Kyuho Bae$^2$, Sung Soo Hwang$^{1,\ast}$, Il Yong Chun$^{3,\ast}$
\thanks{$^\ast$Corresponding Authors}
\thanks{This research was supported by StradVision. We appreciate all the supports of StradVision members who provided insight and expertise. The contents are solely the responsibility of the authors.}}
\address{$^1$School of Computer Science and Electrical Engineering, Handong Global University\\
$^2$Stradvision Inc.\\
$^3$Department of Electrical Engineering, 
University of Hawai`i at M\=anoa
}

\begin{document}

\maketitle

\begin{abstract}

In advanced driver assistant systems and autonomous driving, it is crucial to estimate distances between an ego vehicle and target vehicles.
Existing inter-vehicle distance estimation methods assume that the ego and target vehicles drive on a same ground plane. 
In practical driving environments, however, they may drive on different ground planes.
This paper proposes an inter-vehicle distance estimation framework that can consider slope changes of a road forward, by estimating road gradients of \emph{both} ego vehicle and target vehicles and using a 2D object detection deep net.
Numerical experiments demonstrate that the proposed method significantly improves the distance estimation accuracy and time complexity, compared to deep learning-based depth estimation methods.

\end{abstract}

\begin{keywords}
Inter-vehicle distance estimation, Autonomous driving, ADAS, Visual odometry
\end{keywords}

\section{Introduction}
There have been many advances in advanced driver assistance systems (ADAS) and autonomous driving technologies.
To achieve safe driving, it is important to understand driving environments, such as the existence of obstacles around the vehicle.
Estimating the distances between an ego vehicle to other objects on the road is essential to maintain a safe distance to other vehicles and avoid obstacles, etc.

Many existing inter-vehicle distance estimation methods use LiDAR or RADAR sensors \cite{thakur2016scanning,hakobyan2019high} with the high costs, 
where this paper refers distance estimation between an ego vehicle and target vehicle(s) as inter-vehicle distance estimation.
An alternative to such methods using high-cost sensors is using a monocular camera in distance estimation.
The conventional monocular camera-based distance estimation methods use 2D image processing and physical information of a target vehicle \cite{dagan2004forward, kim2012vision, ali2020real}.
Specifically, the methods need to know the rear width of a target vehicle or the height of the ego vehicle and such requirements limit the practical use of the distance estimation techniques.
Alternatively, deep learning-based depth estimation methods have been proposed \cite{fu2018deep,monodepth2}.
The methods estimate a depth of entire scenes, but does not specifically estimate a distance between vehicles.
These methods have generalization risks, and such risks become problematic if driving environments captured in camera(s) are different between training and test.
In addition, the deep learning-based depth estimation methods are computationally expensive, and it may be challenging to use them combined with object detection deep neural networks (DNNs) in real-time applications, e.g., autonomous driving.
Recently, \cite{zhe2020inter} proposed a camera projection-based inter-vehicle estimation method aided with a 3D object detection DNN.

Camera model-based distance estimation methods assume that ego and target vehicles consistently drive on a same ground plane \cite{zhe2020inter, stein2003vision,gat2005monocular}. 
In practice, however, they may not be on a same ground plane: either ego or target vehicle may drive uphill or downhill, and slope of a road forward may change. 

An alternative is to consider road gradients, where road gradient refers to the steepness of a road.
A flat road is said to have a road gradient of $0^{\circ}$; an uphill is said to have a positive road gradient; a downhill is said to have a negative road gradient. 
The recent distance estimation method \cite{qi2019distance} considers road gradient of an ego vehicle using an inertial measurement unit (IMU). 
Nonetheless, this method does not consider road gradient of a target vehicle, and it might become inaccurate, if road gradient of a target vehicle changes due to slope changes of a road forward.

This paper proposes a distance estimation framework that uses estimated road gradients of both ego vehicle and target vehicles.
To estimate road gradient of an ego vehicle, 
we use the general property that vehicles on a road are parallel to a ground plane and vehicle pose and road gradient changes are closely related.
We estimate ego vehicle's road gradient based on changes of vehicle poses encapsulated in camera rotation matrix at different time points,
where camera rotation matrices are estimated by monocular visual odometry \cite{mur2015orb}.
This is different from \cite{qi2019distance} that uses an IMU in estimating camera rotation matrix.
To estimate road gradient of a target vehicle, we first determine whether ego and target vehicles are driving on a same ground plane.
Specifically, we compare a position of target vehicle estimated by deep-learning based object detection and a vanishing line calculated from the camera rotation matrix estimated above.
Using the comparison results, we estimate a road gradient of a target vehicle.
If ego and target vehicles have sufficiently different road gradients, we adjust road gradient of an ego vehicle by some proper amount.
Numerical experiments with six datasets show that the proposed method significantly improves the distance estimation accuracy and time complexity, compared to deep learning-based depth estimation methods, DORN \cite{fu2018deep} and Monodepth2 \cite{monodepth2}.

\begin{figure}[t]
\centering
\begin{tabular}{c}
  \includegraphics[width=0.7\linewidth]{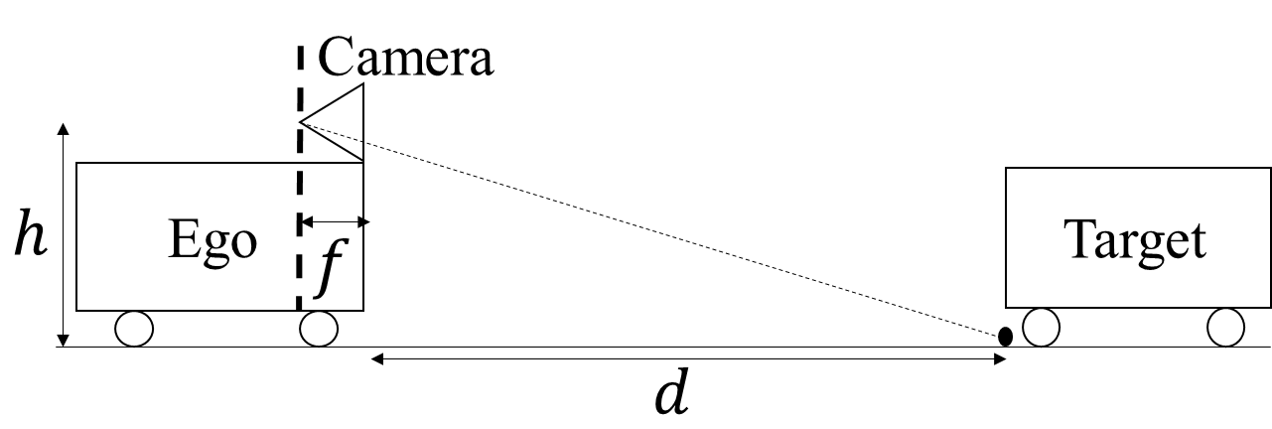}
  \\
  
  (a) Ego and target vehicles are on a same plane.
  \\
  
  \includegraphics[width=0.7\linewidth]{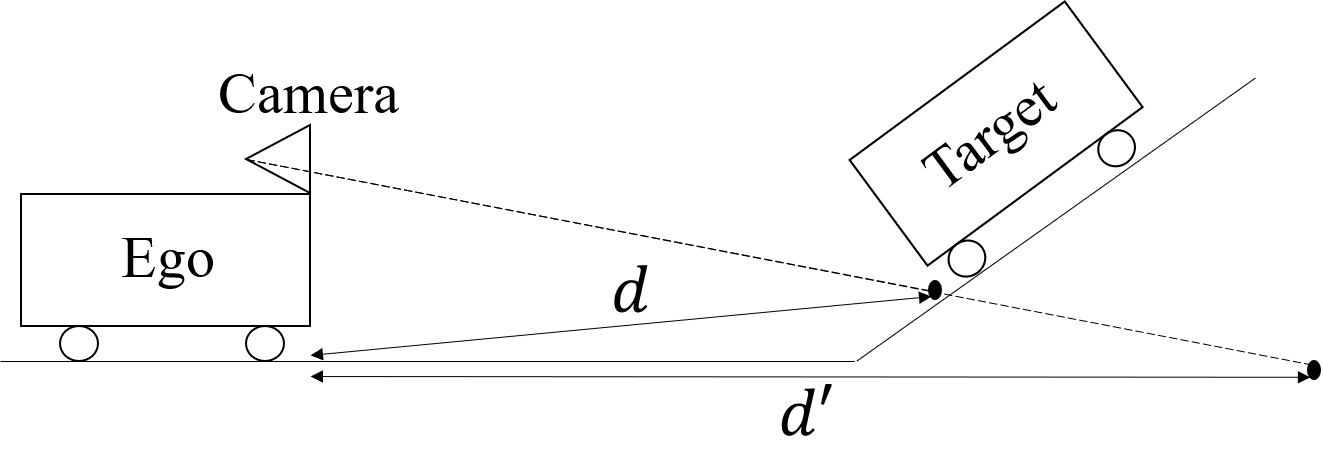}
  \\
  
  (b) Ego and target vehicles are on different planes.
  
\end{tabular}
\vspace{-0.65pc}  
\caption{
Schematic diagrams of distance estimation using a pinhole camera.
}
\vspace{-0.8pc}
\label{fig:schematic_pinhole}
\end{figure}

\section{Background}
This section briefly reviews the pinhole camera model-based distance estimation method. 
The method assumes that ego and target vehicles drive on a same ground plane; Fig.~\ref{fig:schematic_pinhole}(a).
The distance between ego and target vehicles, $d$, can be estimated by the following formula \cite[(3)]{qi2019distance}:
\begin{equation}
\label{eqn:simplified_d}
d = \frac{ f \cdot h }{ \delta_y \cdot (u - v) }
\end{equation}
where $f$ is the camera focal length, 
$h$ is the camera height,
$\delta_y$ is the physical size of a pixel along the $y$-axes in the image domain,
$u$ is a position of the bottom of a target vehicle in an image,
and $v$ is a position of the vanishing line in an image.

The parameters $f$, $h$, and $\delta_y$ are obtained in camera calibration and do not change in driving.
The variable $v$ is affected by a $3 \times 3$ camera rotation matrix $\mb{R}$ \cite[Ch.~8]{hartley_zisserman_2004},
and inaccurate $v$ can cause errors in calculating $d$.
This camera rotation matrix $\mb{R}$ changes according to road changes such as slope and direction changes. 
\cite{qi2019distance} uses an IMU to keep updating $\mb{R}$. 
The variable $u$ also changes while driving, and we observed that object detection DNNs accurately estimate it.

Fig.~\ref{fig:schematic_pinhole}(b) shows a schematic diagram of a pinhole camera when ego and target vehicles drive on different ground planes. 
If a target vehicle's road gradient is not considered, 
method \R{eqn:simplified_d} incorrectly estimates
the distance between ego and target vehicles as $d'$ instead of $d$.

\section{Proposed method}

The proposed method uses video frames and deep learning-based object detection results in estimating road gradients of ego and target vehicles.
We estimate road gradient of an ego vehicle using monocular visual odometry with video frames.
To estimate road gradient of target vehicles, we use a 2D bounding box around a target vehicle that is estimated from deep learning-based object detection. 
To measure a ground plane difference between ego and target vehicles, 
We convert a road gradient of an ego vehicle to a vanishing line,
and compare the position of calculated vanishing line, and that of the center position of a bounding box around target vehicles.
If this measure is less than or equal to some threshold, 
we assume that ego and target vehicles drive on different ground planes; otherwise, they drive on a same ground plane.
If we decide that they drive on different ground planes, 
we adjust road gradient of an ego vehicle and use adjusted one to estimate distance between ego and target vehicles.
If we decide that they drive on a same ground plane, 
road gradient of an ego vehicle is used to estimate inter-vehicle distances.
For simplicity, the section describes the proposed distance estimation method between an ego vehicle and a target vehicle.

\begin{figure}[t]
\centering
\includegraphics[width=0.75\linewidth]{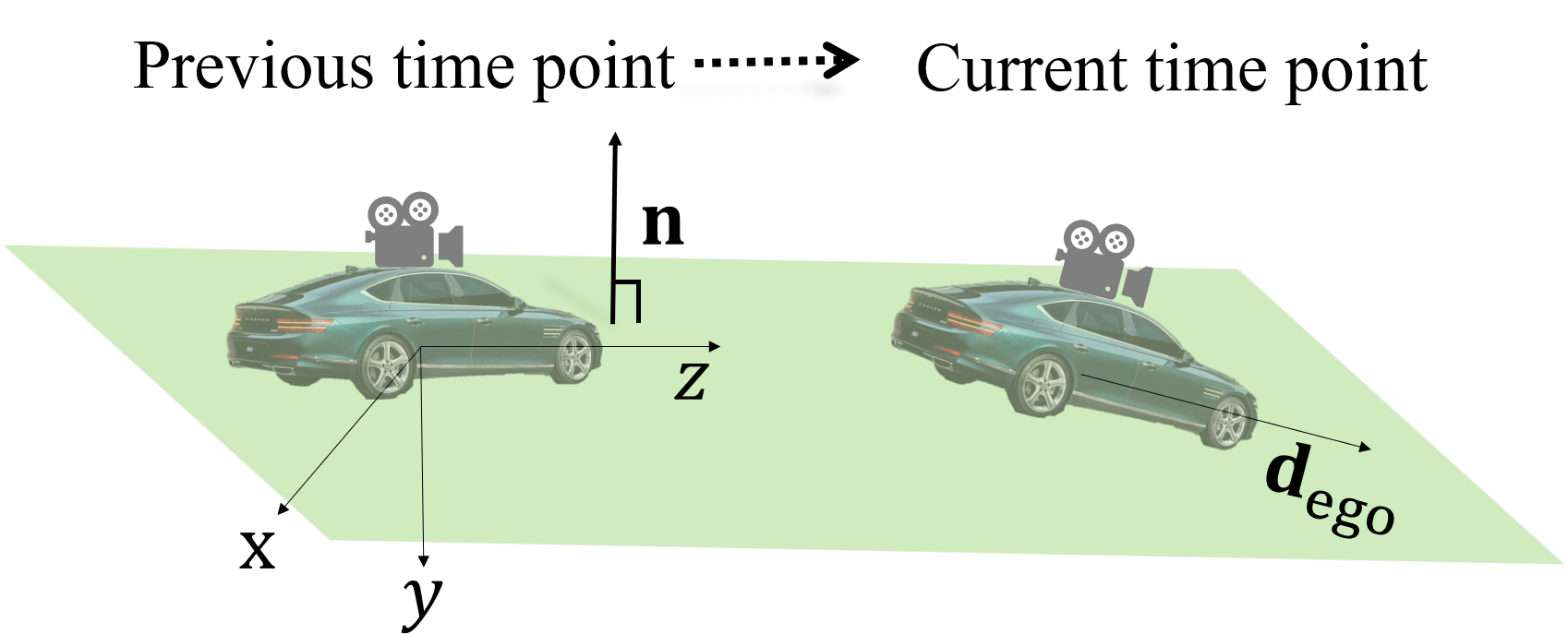}
\vspace{-1pc}  
\caption{Changes of an ego vehicle's pose between previous and current time points.}
\label{fig:cam_axis}
\vspace{-1pc}
\end{figure}

\subsection{Road gradient estimation of an ego vehicle}
\label{sec:ego_road_gradient_estimation}
We estimate road gradient of an ego vehicle using monocular visual odometry.
The monocular visual odometry is defined on the $x$-, $y$- and $z$-axes \cite{{mur2015orb}} -- see Fig.~\ref{fig:cam_axis} -- and calculates a camera rotation matrix difference $\mb{\Delta R} \in \bbR^{3 \times 3}$ between each frame and a starting point.
We assume that at a current time point, an ego vehicle drives in the direction of an unit vector $\mb{z}$.
The proposed method calculates road gradient of an ego vehicle, $\theta$, by calculating an angle between the normal vector $\mb{n}$ to the ground plane, i.e., $xz$-plane in Fig.~\ref{fig:cam_axis},
that is updated at a previous time point, and current direction $\mb{z}$.

Using calculated $\mb{\Delta R}$ via the monocular visual odometry, we calculate the normal vector to the $xz$-plane at a previous time point as follows:
\begin{equation}
\mb{n} 
= \mb{\Delta R}_{\text{prev}} \cdot \mb{n}_0, 
\quad
\mb{n}_0 
= \begin{bmatrix} 0 \\ 
-\cos (\theta_{0}) \\
-\sin (\theta_{0}) 
\end{bmatrix},
\label{eqn:update_normal}
\end{equation}
where
$\mb{\Delta R}_{\text{prev}}$ is the camera rotation matrix difference between previous and initial time points, $\theta_0$ is a camera angle in the $yz$-plane.
We calculate the current direction of an ego by
\begin{equation}
\mb{d}_{\text{ego}} 
= \mb{\Delta R}_{\text{curr}}
\cdot
\mb{d}_{\text{ego},0},
\quad
\mb{d}_{\text{ego},0} 
= \begin{bmatrix} 0 \\ 
0 \\
1 
\end{bmatrix},
\label{eqn:update_dir}
\end{equation}
where 
$\mb{\Delta R}_{\text{curr}}$ is the camera rotation matrix difference between current and initial time points,
$\mb{d}_{\text{ego},0}$ is the default direction of an ego vehicle.
Using \R{eqn:update_normal} and \R{eqn:update_dir}, we calculate road gradient of an ego vehicle by finding an angle between $\mb{n}$ and $\mb{d}_{\text{ego}}$:
\begin{equation}
\theta = 
\frac{\pi}{2} - \arccos{ ( \mb{n}^\top  \mb{d}_{\text{ego}} ) }.
\label{eqn:road_grad}
\end{equation}
noting that both $\mb{n}$ and $\mb{d}_{\text{ego}}$ are unit vectors.

To properly calculate the road gradient of an ego vehicle at a current time point, 
there must be some time interval between previous and current time points. 
If there exist no time interval, then $\mb{\Delta R}_{\text{prev}} = \mb{\Delta R}_{\text{curr}}$ and thus, $\theta = 0$ does not change over time, assuming that $\theta_0 = 0$ (i.e., no camera rotation in the $yz$-plane).
We assume that an ego vehicle drives at $60$ km/h, and set the time interval between previous and current time points as $1$ sec.

\subsection{Road gradient estimation of a target vehicle}
\label{sec:target_road_gradient_estimation}

To estimate road gradient of a target vehicle, 
we use road gradient of an ego vehicle, $\theta$ in \R{eqn:road_grad}, and a 2D bounding box around a target vehicle that is obtained via deep learning-based object detection.
The variable $v$ in \R{eqn:simplified_d} can be calculated with $\theta$ in \R{eqn:road_grad} using the following formula \cite{kovsecka2002video}:
\begin{equation}
\label{eqn:v}
v = c_y - \tan \theta \cdot f,
\end{equation}
where $c_y$ is a $y$-coordinate of the principal point or image center in the image domain (we set the most left corner pixel location as $(0,0)$), and $f$ is given in \R{eqn:simplified_d}. The parameter $c_y$ is obtained in camera calibration and does not change in driving.

\begin{figure}[t!]
\centering
\begin{tabular}{c}
  \includegraphics[width=0.8\linewidth]{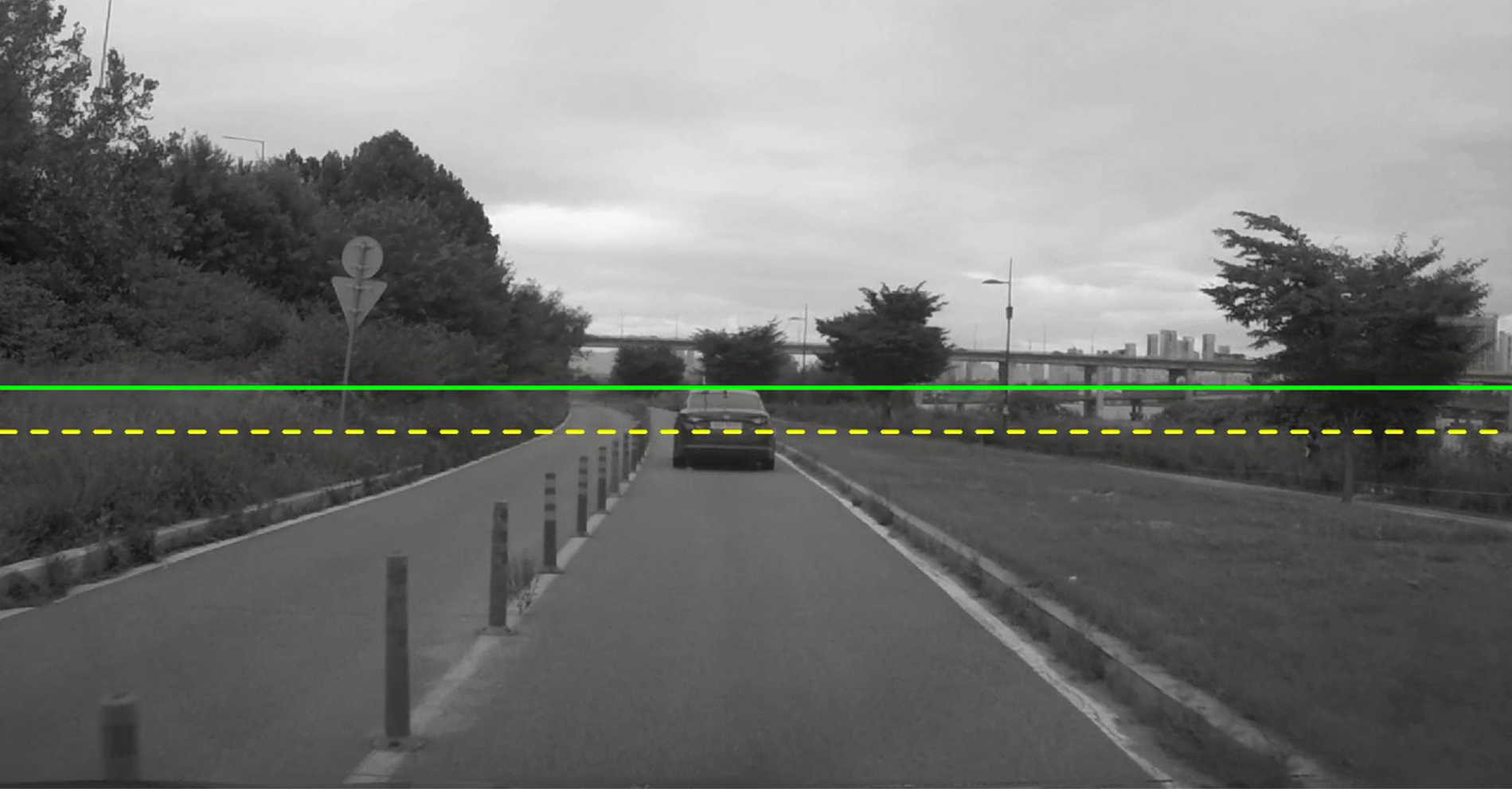}
  \\
  
  (a) Ego and target vehicles drive on a same ground plane.
  \\
  
  \includegraphics[width=0.8\linewidth]{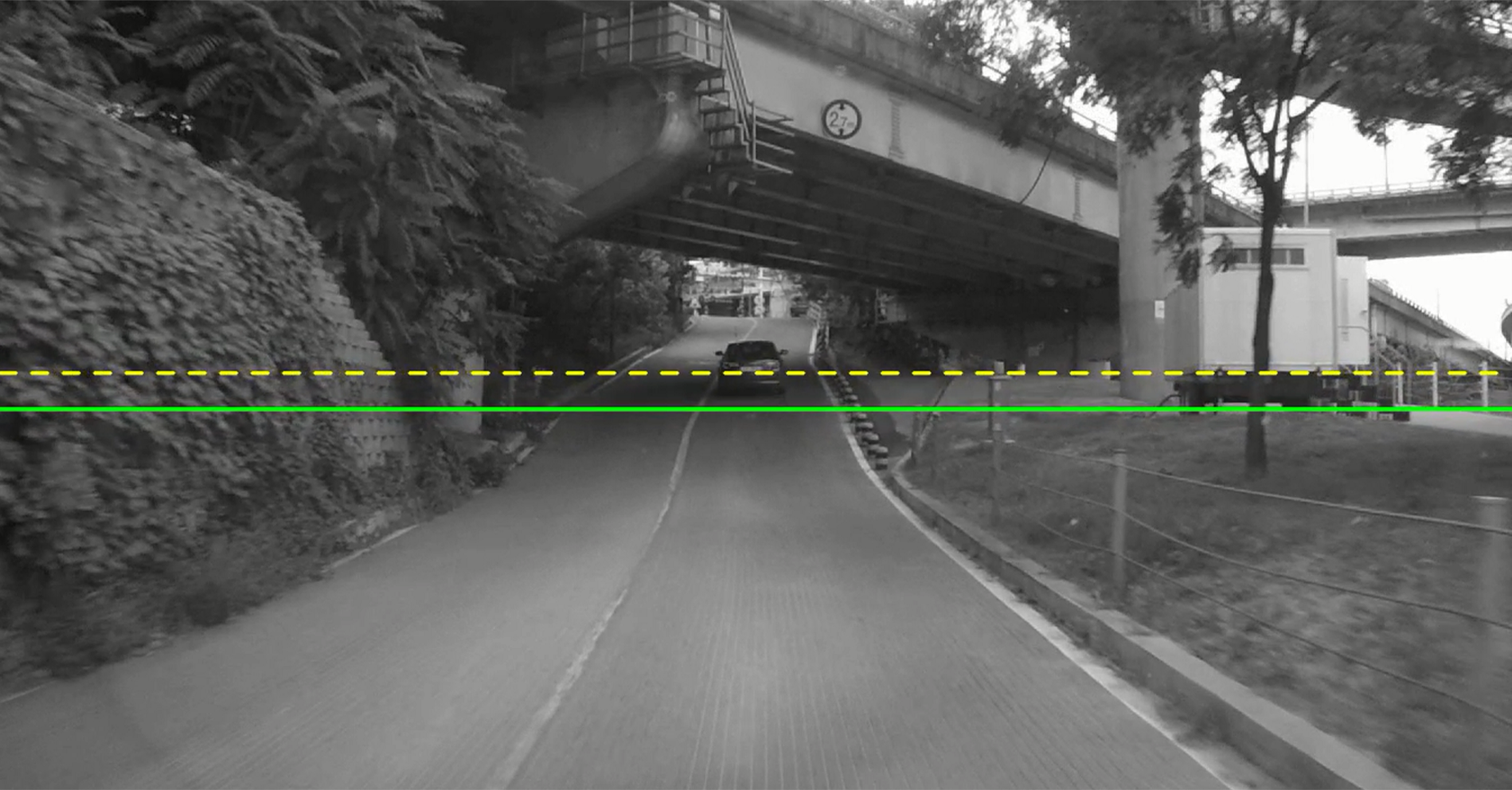}
  \\
  
  (b) Ego and target vehicles drive on different ground planes.
  
\end{tabular}
  
\vspace{-0.5pc}
\caption{
Positions of estimated vanishing line and center of a bounding box around a target vehicle.
Green solid lines denote the vanishing line of a driving scene, and yellow dashed lines denote the center position of 2D bounding box around a target vehicle.
}
\label{fig:case_of_planes}
\vspace{-1pc}
\end{figure}

We observed that when ego and target vehicles drive on different ground planes,
the (bounding box) center of a target vehicle is formed in a deviated location from its usual position formed by those driving on a same ground plane.
When ego and target vehicles drive on a same ground plane, the center of a target vehicle is positioned slightly below a vanishing line; see Fig.~\ref{fig:case_of_planes}(a).
When a target vehicle drives on a low uphill slope, the center of a target vehicle gets closer to a vanishing line.
When a target vehicle drives on a steep uphill slope, the center of a target vehicle can be positioned above the vanishing line; see Fig.~\ref{fig:case_of_planes}(b).
In case of a target vehicle driving on a downhill slope, the center of a target vehicle moves downward from the vanishing line.

We estimate road gradient changes of a target vehicle and adjust road gradient of an ego vehicle by 
using the aforementioned positional relationship between the vanishing line and the center of a target vehicle. 
To estimate road gradient changes of a target vehicle, we calculate the pixel coordinate difference between the vanishing line position, $v$ in \R{eqn:v}, and the center position of a target vehicle, $b_y$, in a $y$-coordinate: 
\begin{equation}
\label{eqn:th}
\Delta_y = b_y - v.
\end{equation}
When ego and target vehicles drive on a same ground plane, the value of $\Delta_y$ is positive.
When a target vehicle starts driving on uphill slope,
$\Delta_y$ becomes smaller,
and if the slope becomes steep, $\Delta_y$ eventually becomes negative.
We mainly investigate the case that a target vehicle drives on uphill slope
(the majority of frames in this paper include either this case or the case that ego and targets vehicles drive on a same ground plane; see Section~\ref{sec:exp}).
We adjust the road gradient of an ego vehicle $\theta$ by considering ground plane difference between ego and target vehicles -- in Fig.~\ref{fig:schematic_pinhole}(b), to obtain $d$ instead of $d'$:
\begin{equation}
  \theta \leftarrow 
  \begin{cases}
    \theta+\alpha_1, & \Delta_y \leq 0,\\ 
    \theta+\alpha_2, & \Delta_y \leq -10,\\
    \theta+\alpha_3, & \Delta_y < -20,\\
    \theta, & \text{otherwise},
  \end{cases}
\label{eqn:update_theta}
\end{equation}
where
$0 < \alpha_1 < \alpha_2 < \alpha_3$ are tunable angle adjustment parameters.
The otherwise condition in \R{eqn:update_theta} implies that we do not adjust $\theta$ if the road gradient difference between ego and target vehicles is small, i.e., the difference between $d$ and $d'$ in Fig.~\ref{fig:schematic_pinhole}(b) is negligible.

\section{Results and Discussions}

\subsection{Experimental setup}
\label{sec:exp}

We compare the proposed method with DORN \cite{fu2018deep} and Monodepth2 \cite{monodepth2}, depth estimation DNN trained in a supervised and unsupervised way, respectively.
For the comparisons, we use the video the sequence of KITTI validation split with $1024 \times 368$ frame size \cite{geiger2012we} (the validation split has sufficiently long depth estimation video) and five video sequences provided by Stradvision (SV) with $1920 \times 1080$ frame size. \cite{stradvision_2021}; we refer them as SV Sequences \#1, \ldots,\#5.
The KITTI video sequence includes cases that ego and target vehicles drive on uphill, downhill and curved road, and 
they drive on \emph{almost} same ground plane in all cases.
Each SV video sequence has different driving environments:
in SV Sequences \#1 and \#2, ego and target vehicles drive on a same flat road and a same downhill slope, respectively;
in SV Sequence \#3, ego and target vehicles drive and curve on a same flat road;
SV Sequence \#4 includes speed bumps and curved flat roads;
SV Sequence \#5 includes the case that a target vehicle drive on steep uphill slope.
The KITTI and SV video sequences include many target vehicles and a single target vehicle, respectively.
In the KITTI video sequence, the proposed method estimates distance between an ego vehicle and \emph{all} target vehicles.
The number of frames of KITTI and SV Sequences is $840$ and $500$, respectively.
The KITTI video sequence obtains the ground truth inter-vehicle distances with LiDAR data.
The SV video sequences obtain those with differential GPS.

In proposed \R{eqn:update_theta}, we set the angle adjustment parameters as follows:
$\alpha_1 = 3^\circ$, $\alpha_2 = 5^\circ$ and $\alpha_3 = 6^\circ$.
We used the SV object detection software \cite{stradvision_2021} to estimate the 2D bounding box of each target vehicle.
We used pre-trained DORN and Monodepth2 networks that were trained by KITTI dataset \cite{fu2018deep,monodepth2}.
These methods consider estimated depth at the bounding box center of target vehicle(s) (obtained by the SV object detection software) as inter-vehicle distance.
The IMU data does not exist in the KITTI and SV datasets, so comparisons with \cite{qi2019distance} are omitted. 

We evaluated the inter-vehicle distance estimation accuracy by the most conventional error metric in the distance estimation and depth estimation applications, root-mean-squared-error (RMSE in m).
In measuring computing times (in secs.) of the proposed method and DORN \& Monodepth2, we used 
Intel Core I7-8700 CPU with 3.20GHz and 64GB RAM, 
and NVIDIA GeForce GTX 1060 with 6GB, respectively.
The measured execution time of the proposed method additionally includes the visual odometry computing time.
All the methods used the 2D object detection software provided by SV, so we did not measure its computation time.

\begin{table}[t!]
\centering
\caption{RMSE values (m) of different inter-vehicle distance estimation methods}
\vspace{0.5pc}

\begin{tabular}{cccc}
\hline
          & Ours          & DORN & Monodepth2 \\ \hline
KITTI     & \textbf{7.63} & 8.35 & 11.5       \\ \hline
SV Seq.~\#1 & \textbf{1.72} & 8.88 & 5.93       \\ \hline
SV Seq.~\#2 & \textbf{3.49} & 7.22 & 9.77       \\ \hline
SV Seq.~\#3 & \textbf{1.25} & 2.86 & 4.63       \\ \hline
SV Seq.~\#4 & \textbf{4.3}  & 6.57 & 8.7        \\ \hline
SV Seq.~\#5 & \textbf{5.95} & 9.03 & 10.35      \\ \hline
\end{tabular}%
\label{tab:RMSE_comparison}
\vspace{-1pc}
\end{table}

\begin{table}[!t]
\centering
\caption{Averaged processing time (secs.) per frame of different methods}
\vspace{0.5pc}

\begin{tabular}{cccc}
\hline
      & Ours          & DORN & Monodepth2 \\ \hline
KITTI & \textbf{0.03} & 3    & 0.06       \\ \hline
SV Seq.~\#1--5    & \textbf{0.04} & 12   & 0.36       \\ \hline
\end{tabular}%
\vspace{-0.5pc}
\label{tab:process_time}
\end{table}

\subsection{Result and discussion}

In all experiments, the proposed method significantly improves the inter-vehicle distance estimation accuracy compared to the deep-learning based depth estimation methods, DORN \cite{fu2018deep} and Monodepth2 \cite{monodepth2}. 
See Table~\ref{tab:RMSE_comparison}.
The SV Sequence \#1--\#3 results in Table~\ref{tab:RMSE_comparison}
show that when ego and target vehicles drive on a same ground plane,
the proposed method significantly improves 
the inter-vehicle distance estimation of an ego vehicle compared to DORN and Monodepth2, regardless of driving environments.
The SV Sequence \#4--\#5 results in Table~\ref{tab:RMSE_comparison} show that when ego and target vehicles drive on different ground planes,
the proposed method significantly improves 
the inter-vehicle distance estimation compared to DORN and Monodepth2.
Before updating road gradients of an ego vehicle (see Section~\ref{sec:target_road_gradient_estimation}),
RMSE values are 9.6 (m) and 13.86 (m) for SV Sequences \#4 and \#5, respectively.
Comparing the results with those in Table~\ref{tab:RMSE_comparison} implies that the proposed method in Section~\ref{sec:target_road_gradient_estimation} is crucial to accurately estimate inter-vehicle distances when ego and target vehicles drive on different ground planes.
Experiments with the KITTI dataset has higher errors than other datasets, because all methods estimate distance between an ego vehicle and many target vehicles.

Table~\ref{tab:process_time} shows that in all experiments, the proposed method is consistently faster than depth estimation DNNs, DORN and Monodepth2.
We expect that the proposed method becomes more efficient than the other depth estimation methods as the resolution of video increases.

\section{Conclusion}

The proposed method achieves accurate and fast inter-vehicle distance estimation by estimating road gradient of \emph{both} ego and target vehicles.
The proposed method is expected to be useful for autonomous driving and ADAS in practical driving environments, where road gradients of ego and target vehicles change over time.

We will investigate an advanced DNN that updates ego vehicle's road gradient $\theta$ based on $v$ in \R{eqn:v} and $b_y$ in \R{eqn:th}.

\bibliographystyle{IEEEbib}
\bibliography{myref}

\end{document}